\documentclass[runningheads]{llncs}
\usepackage{graphicx}
\usepackage[colorlinks, linkcolor=red,anchorcolor=green,citecolor=green]{hyperref}

\usepackage{tikz}
\usepackage{comment}
\usepackage{amsmath,amssymb} 
\usepackage{color}
\usepackage{amssymb}
\usepackage{booktabs}

\usepackage{bbding}

\begin{document}
\pagestyle{headings}
\mainmatter
\def\ECCVSubNumber{690}  

\title{PTSEFormer: Progressive Temporal-Spatial Enhanced TransFormer Towards Video Object Detection} 

\titlerunning{PTSEFormer}
%
\author{Han Wang\inst{1} \and
Jun Tang\inst{2} \and Xiaodong Liu\inst{2} \and Shanyan Guan\inst{3} \and Rong Xie\inst{1} \and Li Song\inst{1,3}\thanks{Corresponding author.}}
\authorrunning{H. Wang et al.}
%
\institute{
Institute of Image Communication and Network Engineering, Shanghai Jiao Tong
University \and
HIKVISION Inc.
\\ \and
MoE Key Lab of Artificial Intelligence, AI Institute, Shanghai Jiao Tong University\\
}

\newcommand{\myparagraph}[1]{\noindent\textbf{#1}.}
\newcommand{\model}[1]{PTSEFormer}

\renewcommand{\thefootnote}{}

\maketitle

\begin{abstract}
Recent years have witnessed a trend of applying context frames to boost the performance of object detection as video object detection. Existing methods usually aggregate features at one stroke to enhance the feature. These methods, however, usually lack spatial information from neighboring frames and suffer from insufficient feature aggregation. To address the issues, we perform a progressive way to introduce both temporal information and spatial information for an integrated enhancement. The temporal information is introduced by the temporal feature aggregation model (TFAM), by conducting an attention mechanism between the context frames and the target frame (\textit{i.e.}, the frame to be detected). Meanwhile, we employ a Spatial Transition Awareness Model (STAM) to convey the 
location transition information between each context frame and target frame. Built upon a transformer-based detector DETR, our PTSEFormer also follows an end-to-end fashion to avoid heavy post-processing procedures while achieving 88.1\% mAP on the ImageNet VID dataset. Codes are available at \url{https://github.com/Hon-Wong/PTSEFormer}.

\end{abstract}

\section{Introduction}

\begin{figure}[t]
    \centering
    \includegraphics[width=\linewidth]{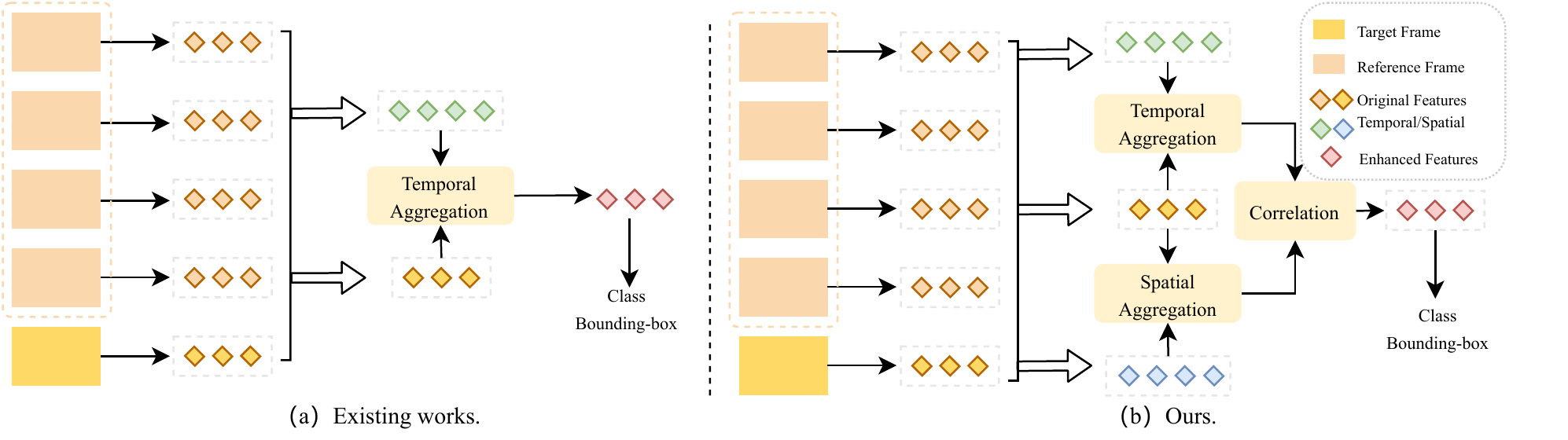}
    \caption{The differences between existing works and ours. Previous works usually conduct temporal feature aggregation at one stroke, lacking in spatial information and suffering from insufficient feature aggregation. In contrast, our PTSEFormer utilizes both spatial and temporal information and performs feature aggregation in a progressive way.}
    
    \label{motivation}
\end{figure}

Video Object Detection (VOD)~\cite{review,fman,fgfa,transvod,mega} has emerged as a hot topic in computer vision. 
Given a target frame and its context frames, VOD aims to detect objects in the target frame, with the compensation of observation from context frames. 
By observing the same instance in different poses from context frames, many hard cases, such as blurry appearance and background occlusion, are possible to be tackled. 

Previous works~\cite{mega,hvrnet,troi,transvod} usually aggregate features at one stroke, suffering from insufficient utilization of temporal information. In particular, they employ isolated box-level associations~\cite{mega,hvrnet,troi} to enhance the instance feature of the target frame only using the extracted features of proposals, ignoring the spatial relations between frames.
To diversify context frame features, those works put effort into how to excavate information from long-range context frames. However, as a common sense of human vision, information from a nearby time window is enough for detection in most scenarios. 
Specifically, when distinguishing a blurry object from the target frame, we often refer to the frame sliding near the target frame temporally, instead of observing the whole video. In this way, how to fully utilize the information from context frames, rather than enlarging the range of context frames, should be valued in the first place. 

In this paper, we propose PTSEFormer to tackle the problems mentioned above. 
Motivated by DETR~\cite{detr,ddetr}, PTSEFormer uses Transformer~\cite{transformer} as the basic structure to avoid complicated post-processing (\textit{e.g.,} Seq-NMS~\cite{seqnms}, Tublet-Linking~\cite{tcnn}, Viterbi~\cite{dt}, Tublet-Rescore~\cite{stlattice}).
In contrast to aggregating features of the target frame and context frames at one stroke by attention layers~\cite{mega,hvrnet,transvod} and conducting box-level associations upon extracted proposals~\cite{mega,hvrnet,troi}, PTSEFormer conducts a progressive way to focus on both the temporal information and the spatial transition relations between frames.
Specifically, \textbf{Temporal Feature Aggregation Module} is designed to introduce the temporal information to enhance the feature of the target frame with different perspectives towards the same objects in all the context frames. 
\textbf{Spatial Transition Awareness Module} is designed for estimating the position transition of the objects between the target frame and each context frame, enhancing the target feature with frame-to-frame spatial information. 
To build a balanced correlation model upon transformer decoder, we further propose the Gated Correlation model, which considers the imbalance caused by the residual connection layer and adds a gate to fix it.

Furthermore, as an important design of DETR, object queries contain inherent object position distribution learned from training data, and are fixed during inferring. We propose the Query Assembling Module(QAM) to regress object queries directly from context frames. Due to the fact that it is more reasonable to infer position from adjacent context frames, rather than from fixed parameters decided by training data.
%
%

We conduct extensive experiments on ImageNet VID dataset~\cite{imagenet} and achieve a \textbf{4.9\%} absolute improvement on mAP compared to previous end-to-end state-of-the-art method~\cite{hvrnet} and \textbf{3.3\%} absolute improvement on mAP compared to its variant with post-processing when applied on a ResNet-101 backbone, showing the effectiveness of our method.

\section{Related Works}

\subsection{Vision Transformer}
Recent years have witnessed great progress on vision transformers. ViT~\cite{vit} first introduces a transformer architecture to the image classification and draws much attention. DETR~\cite{detr,ddetr} builds a transformer-based architecture for object detection, with delicately designed object queries to learn the position distribution of objects. After successful applications, transformers have achieved leading performance in many downstream tasks of computer vision. For instance, in visual object tracking (VOT), TrDimp/TrSiam~\cite{trdimp} modifies the transformer decoder for correlation between features from images, as a replacement of classical correlation model (\textit{i.e.}, depth-wise cross correlation~\cite{siamrpnpp}) in VOT. HiFT~\cite{hift} also utilizes the transformer decoder for correlation on hierarchical features extracted from images via a CNN backbone. The multi-head attentions in the decoder seems naturally suitable for feature correlation. However, we cast doubt on the direct usage of the decoder as a feature fusion model for features in the same feature space. 

\subsection{Video Object Detection}
Object detection suffers from image deterioration problems, such as motion blur, background occlusion, deformation, etc.
To tackle this problem, many works~\cite{longrange,mega,troi,hvrnet} explored to use temporal context frames to provide compensation guidance (\textit{i.e.}, the object at context frames with different viewpoints). 
Built upon a two-stage detector (\textit{e.g.}, Faster-RCNN~\cite{fasterrcnn}, R-FCN~\cite{rfcn}, FPN~\cite{fpn}), Early works~\cite{mega,hvrnet,troi,SELSA} conduct box-level associations and achieve remarkable success. 
However, these methods highly rely on the features of proposals extracted by the two-stage detector, lacking spatial information. 
In recent years, the rapid progress of anchor-free object detectors obtain remarkable performance. 
We observe several attempts to introduce anchor-free methods to video object detection and boost the performance by spatial information. 
CHP~\cite{chp} uses an anchor-free detector CenterNet~\cite{centernet} as a base detector and propagates its heat map by post-processing to deliver the spatial information. Apparently, it ignores the support from the temporal features.
TransVOD~\cite{transvod} is the first to apply transformer architecture~\cite{transformer} into VOD and builds upon DETR. 
However, suffering from insufficient feature aggregation and lacking spatial information, its performance is inferior to those with box-level associations when applied on the same backbone.
%
To address the limitations mentioned above, we propose an end-to-end framework with temporal-spatial feature aggregation design to better employ context frames information. 

\begin{figure}[t]
    \centering
    \includegraphics[width=\linewidth]{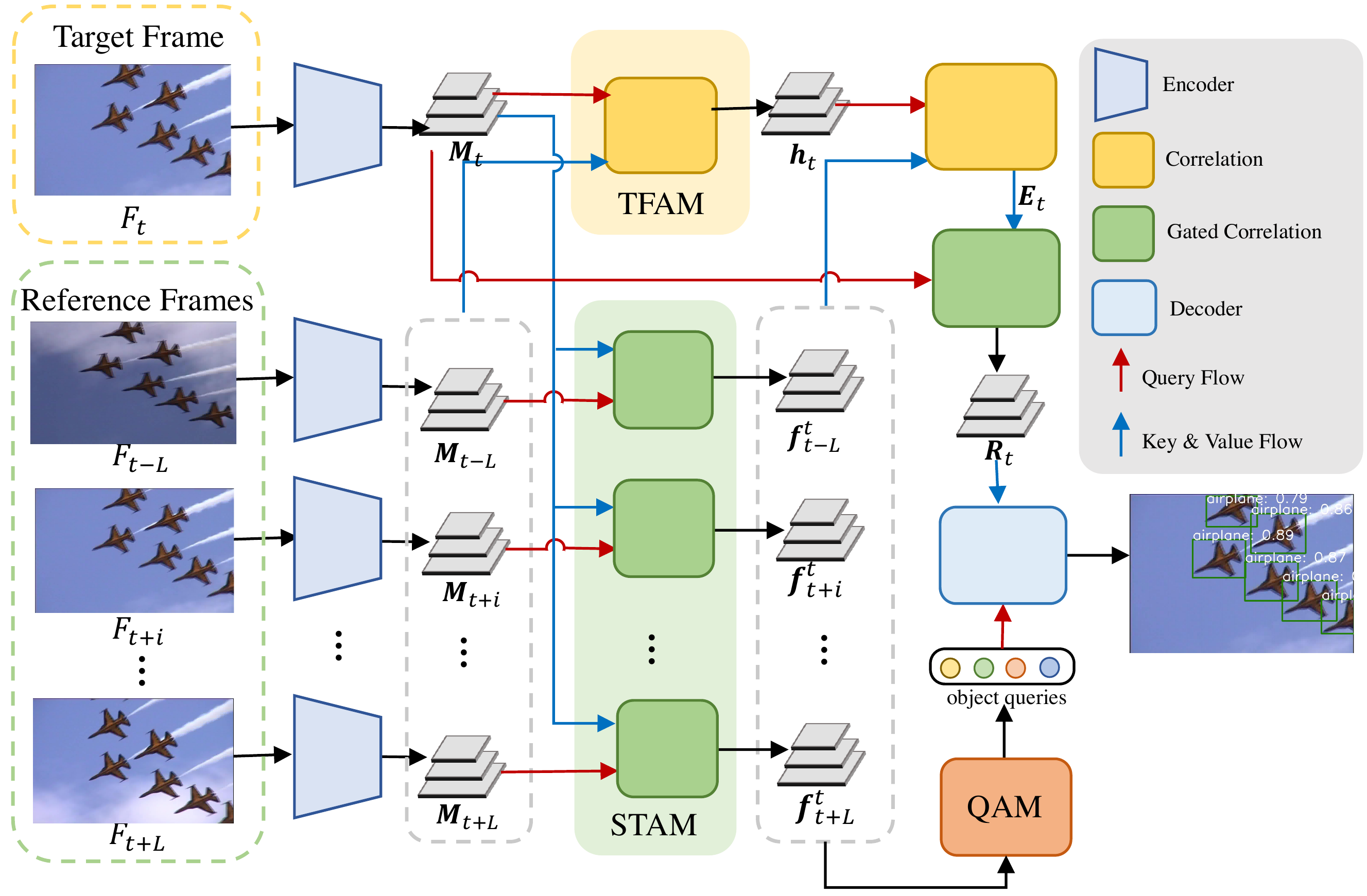}
    \caption{Overview of the proposed \model{}. First, image features $\boldsymbol{M}$ are extracted by a transformer-based encoder. The image features are further input to TFAM and STAM to obtain temporal feature $\boldsymbol{h}_t$ and spatial features $\{\boldsymbol{f}_t^i\}_{i=-L:L}$, and then are progressively aggregated. Finally, the aggregated feature, together with regressed object queries from QAM, is decoded for final detection result}

    
    \label{framework}
\end{figure}

\section{PTSEFormer}

\subsection{Overview}
The overview of PTSEFormer is shown in Figure~\ref{framework}.
Given a target frame $F_t$ and its context frames $F_t^c = \{F_{t+i}\}_{i=-L:L}$, \model{} detects the class and bounding-box of objects at $F_t$.
To better explore the context information from $F_t^c$, \model{} extracts both temporal features (representing the motion of objects) and spatial features (representing position and transformations of objects). 
Next, the temporal and spatial features are progressively aggregated. Then a decoder learns to infer the class and bounding boxes from the aggregated feature and the object query. 
Particularly, our object query is conditioned on $F_t^c$, and thus leads to more accurate object position distribution. 

In Section~\ref{sec:TFAM_STAM}, we introduce the details of encoding temporal and spatial memories, including feature extraction and progressive aggregation. Next, Section~\ref{sec:decoding} introduces how to infer class and bounding-box of objects from the aggregated feature. Finally, the details of learning \model{} are described in Section~\ref{sec:learning}, including the total objective function and the network details.

\subsection{Temporal and Spatial Encoding}
\label{sec:TFAM_STAM}
We introduce how to extract temporal and spatial memories from the target frame $F_t$ and its context frames $F_t^c$. First, a transformer-based encoder embeds $F_t$ and $F_t^c$ to latent feature maps respectively, termed as $\boldsymbol{M}_t$ and $\boldsymbol{M}_{t}^c=\{\boldsymbol{M}_{t+i}\}_{i=-L:L}$. Then our model obtains the temporal and spatial memories from $\boldsymbol{M}_t$ and $\boldsymbol{M}_{t}^c$ by two modules: Temporal Feature Aggregation Module (TFAM) and Spatial Transition Awareness Module (STAM). Finally, the temporal and spatial memories are progressively aggregated. We describe the details of each module below.

\myparagraph{TFAM} As demonstrated in previous works~\cite{transvod,mega,hvrnet,longrange}, learning the temporal relation between $F_t$ and $F_t^c$ is beneficial for detecting objects with blurry appearance or distorted shape. Consequently, we propose TFAM to extract this temporal memory $\boldsymbol{h}_t$, which is formulated as:
\begin{align}
    \boldsymbol{h}_t = \mathcal{C}(\boldsymbol{M}_t, \boldsymbol{M}_{t}^c),
\end{align}
where $\mathcal{C}(\cdot, \cdot)$ is the correlation operator:
\begin{align}
    \mathcal{C}(Q, V) = \text{softmax}(\frac{QK^T}{\sqrt{d_k}}) V + Q,
    \label{eq:correlation}
\end{align}
where $Q \in \mathbb{R}^{N_Q \times d_k}$, $K,V \in \mathbb{R}^{N_V \times d_k}$, 
and `$+$' represents the residual connection.


\myparagraph{STAM}
STAM is proposed to learn relative positional transition of objects from a context frame $F_{t+i}$ to the target frame $F_{t}$. 
Since the object identity annotation is unavailable in the VOD task, unsupervised learning of the relations of the objects at $F_{t+i}$ and $F_{t}$ is non-trivial.  

A straightforward idea is to employ the correlation operator $\mathcal{C}(\cdot)$ to model the relative transitions between $F_{t+i}$ and $F_{t}$. 
However, the imbalance weight on $Q$ and $V$ in Equation~\ref{eq:correlation} makes it infeasible to match the objects at two frames. 
Specifically, the weights before $Q$ and $V$ are $1$ and $\text{softmax}(\frac{QK^T}{\sqrt{d_k}})$, respectively. The average value of $\text{softmax}(\frac{QK^T}{\sqrt{d_k}})$ is decided by the size of $Q$ and $K$. When the size goes large, the weight is far less than $1$, leading to severer imbalance attention on $Q$ and $V$. Commonly, this architecture is used for correlation between features from different space and dimensions, which naturally need biased attention. However, in some recent researches~\cite{trdimp,transvod,hift}, it is also used for correlation between features in the same spaces without any modification. We believe the imbalanced attention could do harm to the performance. 

To address the limitation mentioned above and inspired by the gate control design by GRU~\cite{gru}, we design a \textbf{Gated Correlation operation}, denoted as $\mathcal{C}^g$. By adding a gate control to the residual connection of the decoder, we can change the weight before $Q$. Furthermore, to get the gate control awareness of the input $Q$ and $V$, the control weight must be decided by $Q$ and $K$. Thus, we pass $Q$ and $K$ through a fully connected gate layer for the weight. The process can be changed into:
\begin{align}
    \mathcal{C}^g &= \text{softmax}(\frac{QK^T}{\sqrt{d_k}})V+M\odot Q+(1-M)\odot V,\\
    M &= \sigma(\mathcal{G}([Q,V])),
\end{align}
where $\mathcal{G}(\cdot)$ refers to the gated function, consisting of a fully connected function. $\sigma(\cdot)$ is the Sigmoid function. $[\cdot, \cdot]$ is the concatenation operation, and $\odot$ refers to the Hadamard production. Note that $Q$, $K$, $V$ and $M$ must be of the same size. When initializing, the Sigmoid function in gate can project the output to (0, 1) with a primal value of 0.5, conducting fair attention on both $Q$ and $V$.

The final STAM can be formulated as:
\begin{align}
    \boldsymbol{f}^t_{i} = \mathcal{C}^g(\boldsymbol{M}_{t}, \boldsymbol{M}_{t+i}),
\end{align}
where $i=-L:L$, and $\boldsymbol{f}^t_{i}$ is the extracted spatial memory.

\begin{figure}[t]
    \centering
    \includegraphics[width=0.9\linewidth]{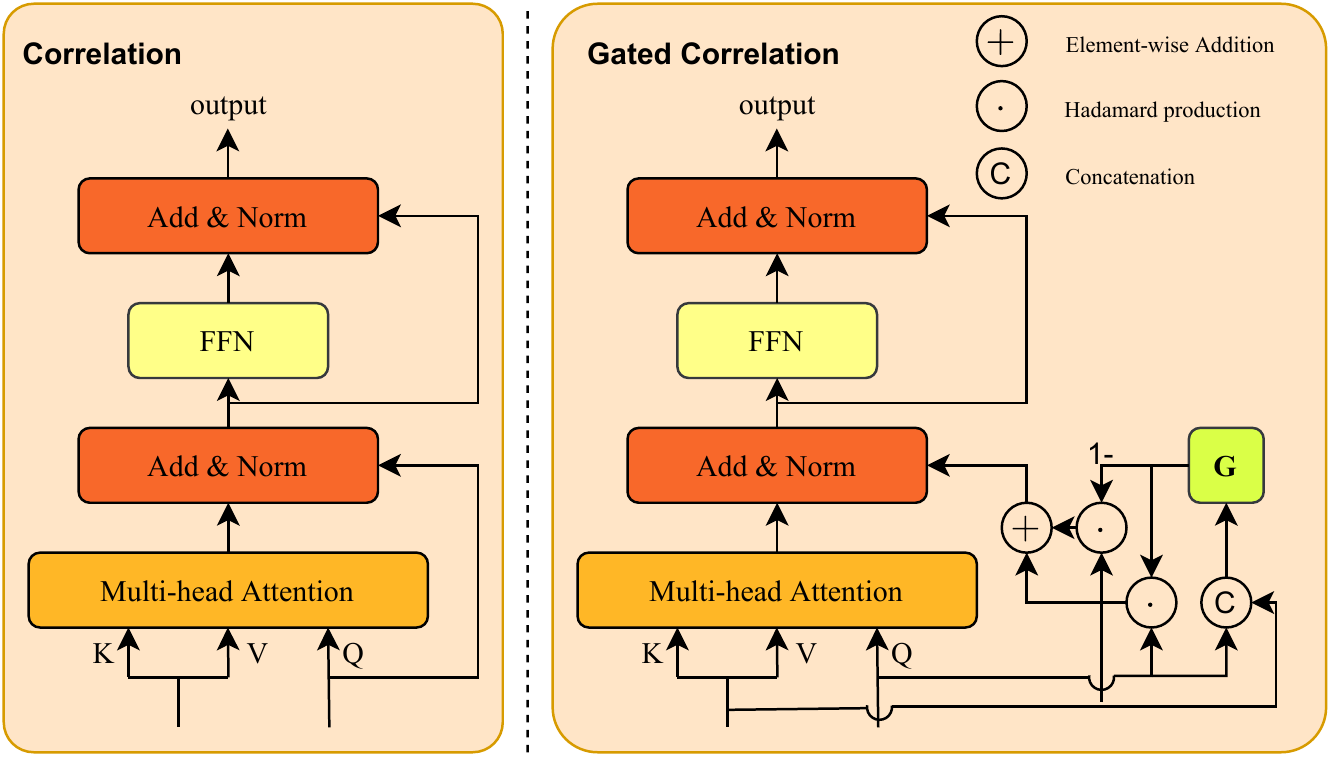}
    \caption{Illustration of Correlation (left) and Gated Correlation (right).}
    \label{fig:corr}
\end{figure}

\myparagraph{Progressive Aggregation} We aggregate the $\boldsymbol{h}_t$ and $\boldsymbol{f}^t_i$ in a progressive way.
First, $\boldsymbol{h}_t$ and $\boldsymbol{f}^t_i$ are combined with the Correlation operation $\mathcal{C}(\cdot)$ to generate a temporal-spatial memory $\boldsymbol{E}_t$. The formulation is written as:
\begin{align}
    \boldsymbol{E}_t = \mathcal{C}(\boldsymbol{h}_t, \{\boldsymbol{f}^t_i\}_{i=-L:L}).
\end{align}
By aggregating features from context frames, $\boldsymbol{E}_t$ contains both long-term temporal and spatial transition information.
However, in some scenes, the context frames are likely to be low-quality and the spatial-temporal memory may be useless and even misleading. In this situation, we should take more information of the current frame instead of context frame. Thus, we use a Gated Correlation between the feature of current frame and the temporal-spatial memory to obtain the final enhanced memory $\boldsymbol{R}_t$. The operation is denoted as Residual Gated Correlation, which can be written as:
\begin{equation}
    \boldsymbol{R}_t = \mathcal{C}^g(\boldsymbol{E}_t, \boldsymbol{M}_t), 
\end{equation}

\subsection{Enhanced Memory Decoding}
\label{sec:decoding}
In original DETR, a group of learned embeddings is designed to learn the position distribution of different objects. With each object query, the decoder decodes one bounding box and its class on the memory. Following the same protocols, we decode our enhanced memory $\boldsymbol{R}_t$ with a transformer decoder. However, there remains a question that the original object queries are fixed through time, cannot benefit from the context frames. Thus, we propose a Query Assembling Model to diversify the object query and convey the position distribution information through time.
\begin{figure}[t]
    \centering
    \includegraphics[width=0.7\linewidth]{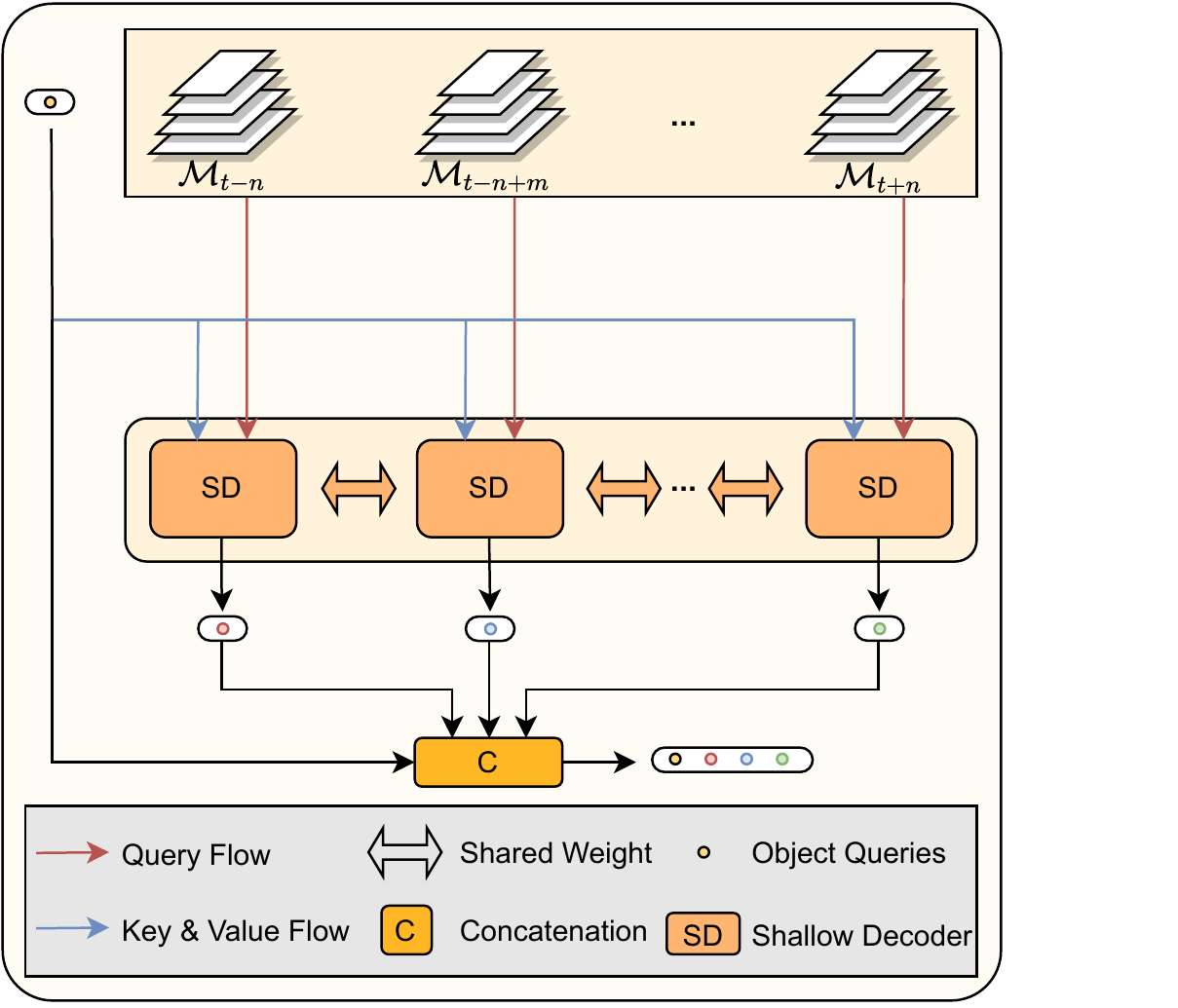}
    \caption{Illustration of the Query Assembling Model (QAM). We apply a shared shallow decoder to combine the primal object queries and each context frame. All the output from shallow decoders are concatenated to form the final object queries.}
    \label{fig:gam}
\end{figure}

\myparagraph{Query Assembling Model} 
Query Assembling Model aims at propagating implicit position distribution information via object queries through time. As primal object queries in DETR are fixed embeddings in the inference stage and have no difference across frames, we apply a shallow correlation model to inherit location information of the primal object queries and diversify information from features. The final object queries can be described as:
\begin{equation}
    {Q} = [{Q}_p, \{\mathrm{SD}(Q_p, \boldsymbol{M}_{t+i}),i=-L:L\}], 
\end{equation}
where ${Q}_p$ is the primal object query, and $\mathrm{SD}$ is a shallow transformer decoder with 2 layers. $[\cdot, \cdot]$ is the concatenation operation.


\subsection{Learning PTSEFormer}
\label{sec:learning}
Following DETR, we adopt a Hungarian algorithm~\cite{hungarian} to calculate the matching cost between the ground truths and predictions. The objective function is formulated as follows:
\begin{align}
    \mathcal{L} &= \lambda_{cls}\mathcal{L}_{cls} + \lambda_{box}\mathcal{L}_{box}, \\
    \mathcal{L}_{box} &= \lambda_{L1}\mathcal{L}_{L1} + \lambda_{giou}\mathcal{L}_{giou},
\end{align}
where $\mathcal{L}_{cls}$ is focal loss~\cite{focal} for classification. $\mathcal{L}_{L1}$, $\mathcal{L}_{giou}$ represent the L1 loss and GIoU loss~\cite{giou} for bounding box regression, respectively. $\lambda_{cls}$, $\lambda_{box}$, $\lambda_{L1}$, $\lambda_{giou}$ are hyper-parameters to balance the multi-task losses.

\myparagraph{Network Details} \model{} is built upon the DETR with several modifications. The number of layers in the encoder and decoder is decreased to 2 for a trade-off between speed and precision. Notice that our method also adopts multi-scale features to boost the performance for detecting small objects. We adopt the ResNet models as our backbones. In particular, we adopt ResNet-101~\cite{resnet} for a fair comparison with previous works. All the components (\textit{i.e.}, TFAM, STAM, Correlation and Gated correlation) also have a two-layer structure. The number of heads in multi-head attention is fixed as 6 and the number of primal object queries is set to be 100, the same as the original DETR.

\section{Experiments}

\subsection{Implement Details}
\myparagraph{Dataset and Metric} 
For a fair and convincing comparison, we conduct our experiments on ImageNet VID dataset~\cite{imagenet} which is a large-scale public dataset for video object detection and contains more than 1M frames for training and more than 100k frames for validation. In particular, we train our model on the training split of ImageNet VID and DET dataset~\cite{imagenet} following common protocols. Same as previous works~\cite{mega,transvod}, we adopt mean average precision (mAP) as our metric.

\myparagraph{Training Details} We train our \model{} on 8 GPUs of Tesla V100 with Adam~\cite{adam}, and each GPU holds one target frame and its reference frames. The whole training procedure lasts for 50 epochs, each taking almost 1.5 hours. The initial learning rate is 1e-4, with a drop in the 40th epoch to 1e-5. For each target frame, we randomly sample 2 frames from a sliding window with a length of 25 as the reference frames. The input images are all resized to hold a shorter size of 800 pixels without any other extra data augmentation applied. All the networks including the single frame baseline are trained from the very beginning with a pre-trained backbone.




\subsection{State-of-the-art Comparison}

\begin{table}[!t]
    \centering
    \setlength{\tabcolsep}{2pt}
    \begin{tabular}{llccc}
    \toprule
        Methods & Base Detector & Stages & Backbone & mAP(\%) \\
        \midrule
        DFF~\cite{dff} & R-FCN & 2 & ResNet-50 & 70.4\\
        FGFA~\cite{fgfa} & R-FCN & 2 & ResNet-50 & 74.0\\
        RDN~\cite{rdn} & Faster-RCNN & 2 & ResNet-50 & 76.7\\
        MEGA~\cite{mega} & Faster-RCNN & 2 & ResNet-50 & 77.3\\
        TransVOD~\cite{transvod} & Deformable DETR & 1 & ResNet-50 & 79.9 \\
        OURS & Deformable DETR & 1 & ResNet-50 & \textbf{87.4} \\
        
    \bottomrule
    \end{tabular}
    \caption{End-to-end methods comparisons (with ResNet-50 backbone).}
    \label{r50}
\end{table}

\begin{table}[!t]
    \centering
    \setlength{\tabcolsep}{2pt}
    \begin{tabular}{llccc}
    \toprule
        Methods & Base Detector & Stages & Backbone & mAP(\%) \\
        \midrule
        LLTR~\cite{longrange} & FPN & 2 & ResNet-101 & 81.0\\
        DFF~\cite{dff} & R-FCN & 2 & ResNet-101 & 73.0 \\
        D\&T~\cite{dt} & R-FCN & 2 & ResNet-101 & 75.8 \\
        LSTS~\cite{lsts} & R-FCN & 2 & ResNet-101 & 77.2 \\
        FGFA~\cite{fgfa} & R-FCN & 2 & ResNet-101 & 76.3 \\
        SELSA~\cite{SELSA} &  Faster-RCNN & 2 & ResNet-101 & 80.3\\
        TROI~\cite{troi} + SELSA~\cite{SELSA} &  Faster-RCNN & 2 & ResNet-101 & 82.0\\
        MEGA~\cite{mega} & Faster-RCNN & 2 & ResNet-101 & 82.9\\
        HVRNet~\cite{hvrnet} & Faster-RCNN & 2 & ResNet-101 & 83.2\\
        \midrule
        CHP~\cite{chp} & CenterNet & 1 & ResNet-101 & 76.7 \\
        TransVOD~\cite{transvod} & Deformable DETR & 1 & ResNet-101 & 81.9 \\
        OURS & Deformable DETR & 1 & ResNet-101 & \textbf{88.1} \\
        
    \bottomrule
    \end{tabular}
    \caption{End-to-end methods comparisons (with ResNet-101 backbone).}
    \label{r101}
\end{table}

\begin{table}[!h]
    \centering
    \setlength{\tabcolsep}{2pt}
    \begin{tabular}{llcclc}
    \toprule
        Methods & Base Detector & Stages & Backbone & Post-processing & mAP(\%) \\
        \midrule
        PSLA~\cite{psla} & R-FCN & 2 & ResNet-101 & Seq-NMS & 81.4 \\
        D\&T~\cite{dt} & R-FCN & 2 & ResNet-101 & Viterbi & 79.8 \\
        MANet~\cite{manet} & R-FCN & 2 & ResNet-101 & Seq-NMS & 80.3 \\
        Scale-Time Lattice~\cite{stlattice} & R-FCN & 2 & ResNet-101 & Tublet-Rescore & 79.6 \\
        FGFA~\cite{fgfa} & R-FCN & 2 & ResNet-101 & Seq-NMS & 78.4 \\
        SELSA~\cite{SELSA} & Faster-RCNN & 2 & ResNet-101 & Seq-NMS & 82.5\\
        MEGA~\cite{mega} & Faster-RCNN & 2 & ResNet-101 & Seq-NMS & 84.5\\
        HVRNet~\cite{hvrnet} & Faster-RCNN & 2 & ResNet-101 & Seq-NMS & 84.8\\
        
        \midrule
        CHP~\cite{SELSA} & CenterNet & 1 & ResNet-101 & Seq-NMS & 78.4 \\
        TransVOD~\cite{transvod} & Deformable DETR & 1 & ResNet-101 & - & 81.9 \\
        OURS & Deformable DETR & 1 & ResNet-101 & - & \textbf{88.1} \\
        
    \bottomrule
    \end{tabular}
    \caption{State-of-the-art methods comparisons (with Post-processing).}
    \label{post}
\end{table}
We first compare our PTSEFomer with several state-of-the-art methods in an end-to-end fashion. As shown in Table \ref{r50} and Table \ref{r101}, we group these methods into two categories by their backbones. Previous end-to-end methods are also mostly built upon a two-stage detector without a post-processing procedure for VOD. 
The existing one-stage based VOD approaches, however, fall behind. Built upon a one-stage detector, we achieve much higher performance on mAP than existing methods with a magnificent margin. Reasonably, the larger backbone boosts the performance of all the methods, including ours. As illustrated in Table \ref{r50} and Table \ref{r101}, Our PTSEFomer leads the performance with ResNet-50 and ResNet-101~\cite{resnet}.

We also compare our PTSEFormer with several state-of-the-art methods with post-processing procedures in Table \ref{post}. Post-processing proves useful in many VOD methods, especially in those built upon an anchor-based detector. Indeed, most existing methods have their versions with post-processing to boost the performance. For instance, the most widely used post-processing, Seq-NMS, conducts an NMS operation through a sequence, boosting the mAP by 1\%-2\%. However, those post-processing procedures, though prove effective, demand extra computations. Thus, our \model{} obtains an end-to-end structure. We declare that even we do not adopt post-processing, our method still obtains the best score on mAP. 

\subsection{Ablation Studies}
Considering the speed, we adopt the ResNet-50 model as our backbone for ablation study. The effectiveness of each component of \model{} is verified independently.
\begin{table}[!t]
    \centering
    \setlength{\tabcolsep}{2pt}
    \begin{tabular}{lccc}
    \toprule
        Method&STAM&TFAM&mAP(\%) \\
        \midrule
        Single Frame Baseline~\cite{ddetr} &\XSolidBrush  &\XSolidBrush  &  81.2\\
        \model{} &  \Checkmark & \XSolidBrush & 84.5\\
        \model{} & \Checkmark & \Checkmark & \textbf{87.4}\\
    \bottomrule
    \end{tabular}
    \caption{Ablation studies of STAM and TFAM.}
    \label{ablation1}
\end{table}

\begin{table}[!t]
    \centering
    \setlength{\tabcolsep}{2pt}
    \begin{tabular}{ccccc}
    \toprule
       QAM&GatedCorr&RGC&Multi-scale&mAP(\%) \\
        \midrule
         \XSolidBrush & \Checkmark & \Checkmark & \Checkmark & 86.1\\
          \Checkmark & \XSolidBrush & \Checkmark & \Checkmark & 86.7\\
          \Checkmark & \Checkmark & \XSolidBrush & \Checkmark & 86.3\\
        \Checkmark &\Checkmark &\Checkmark &\XSolidBrush & 86.4\\
          \Checkmark & \Checkmark & \Checkmark & \Checkmark & \textbf{87.4}\\
    \bottomrule
    \end{tabular}
    \caption{Ablation studies on QAM, Gated Correlation, RGC and Multi-scale.}
    \label{ablation2}
\end{table}


\myparagraph{TFAM and STAM} To verify the effectiveness of the TFAM and STAM, we conduct ablation studies on both, respectively.  As shown in Table \ref{ablation1}, we add our STAM model and TFAM model step by step to verify the effectiveness of both. The use of STAM improves the mAP by 3.3\%, performing spatial relations between the target frame and each reference frame and offering spatial transferring information. As mentioned above, TFAM conducts a temporal feature aggregation, providing the temporal memory of the target frame. The TFAM leads to an increase of 2.9\% compared with only applying STAM.


\myparagraph{Query Assembling Model} Query Assembling Model carries the spatial information through time, offering implicit track information. The original object queries in DETR are fixed embeddings, expected to learn the position distribution of the objects in the dataset. We compare QAM with the original object queries in DETR in our experiment by replacing the QAM with original object queries. By comparing line 1 and line 5 in Table \ref{ablation2}, results have shown the assistance from the specially designed QAM by an improvement of 1.3\% on mAP.

\myparagraph{Gated Correlation} To alleviate the imbalanced attention on Key and Value of the transformer decoder as a correlation model, we propose Gated Correlation to carry out a relation between temporal memory and spatial memories. To prove it useful, we replace it with the original transformer decoder. The results show a little drop in mAP which is illustrated in line 2 and line 5 in Table \ref{ablation2}.

\myparagraph{Residual Gated Correlation} The Residual Gated Correlation model is designed for gating out the memories from low-quality reference frames and boosts the performance of our method. We also investigate it in our experiment and the results from line 3 and line 5 in Table \ref{ablation2} show its positive influence on the performance. In particular, application of Residual Gated Correlation leads to a 1.1\% increasement on mAP.

\myparagraph{Multi-scale} Similar to the original DETR, the designs of our methods also benefit from the multi-scale features. We obtain 1\% increment on mAP with a multi-scale architecture by comparing line 4 and line 5 in Table \ref{ablation2}.


\subsection{Visualization}
\begin{figure}[!ht]
    \centering
    \includegraphics[width=\linewidth]{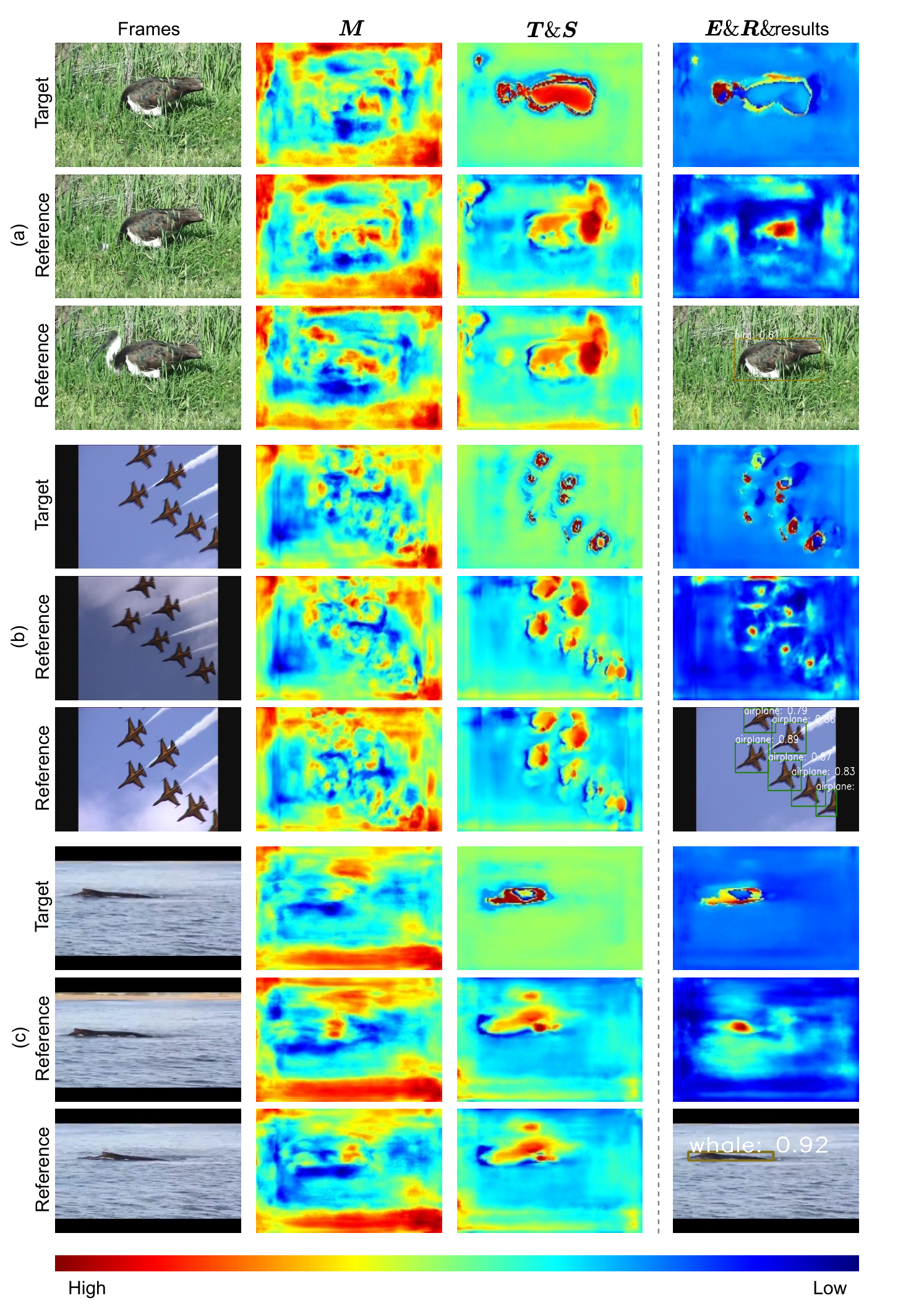}
    \caption{The feature maps of our models. We select three target frames to figure out what the network learns. }
    \label{fig:attention}
\end{figure}

\begin{figure}[!ht]
    \centering
    \includegraphics[width=\linewidth]{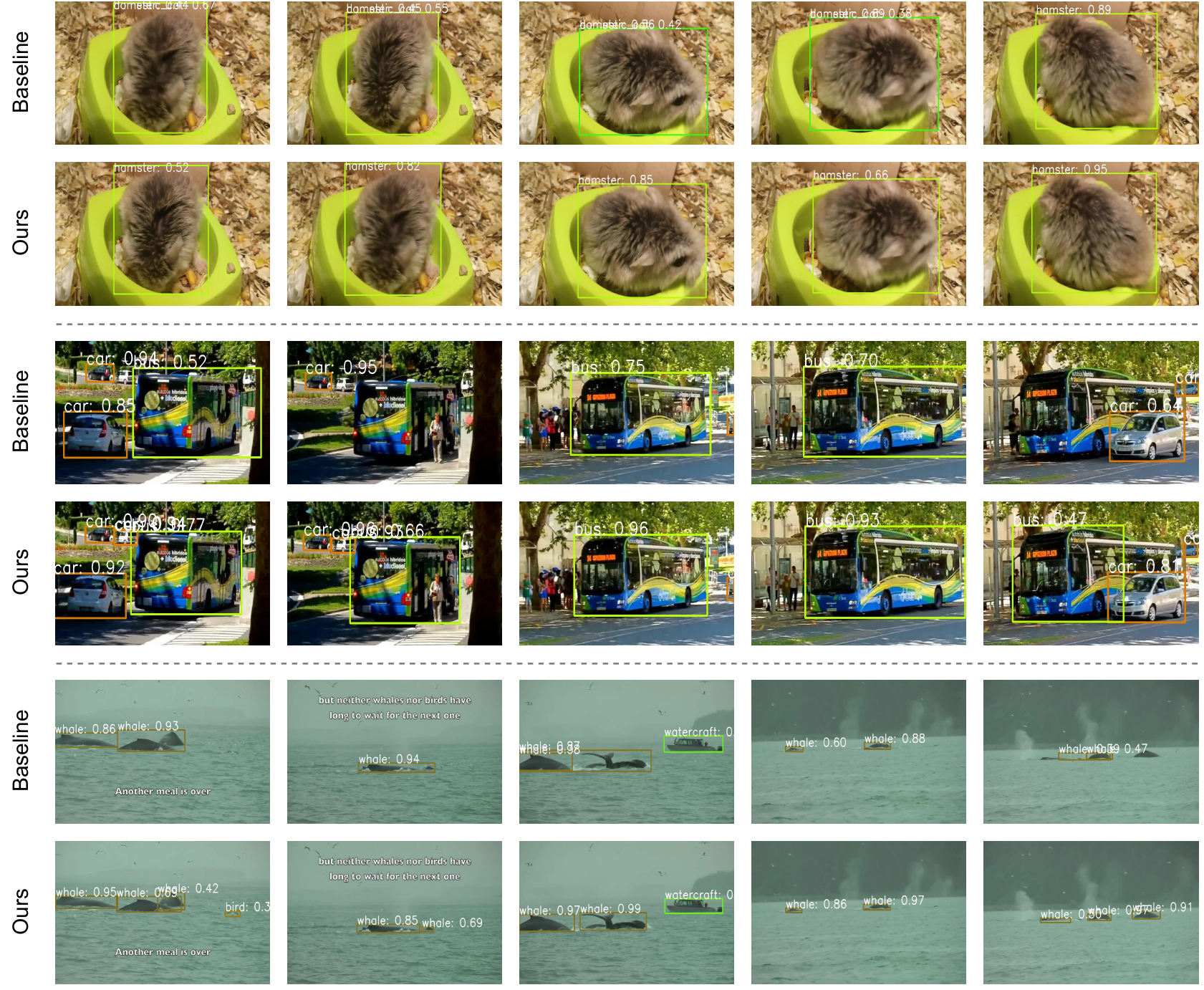}
    \caption{Results Visualization.  Our results are in the odd row, and single frame detector DETR results as baseline are in the even row. As shown in figure, our method is more robust against various image deterioration(\textit{e.g.},occlusion, deformation).
    %
    }
    \label{fig:samples}
\end{figure}
\myparagraph{Feature Visualization} We first visualize the feature maps of our network to figure out how our TFAM and STAM work. As depicted in Figure \ref{fig:attention}, we demonstrate three target frames and their corresponding reference frames and feature maps, respectively. The first column shows the original input frames (\textit{i.e.}, target frame and its two reference frames, from top to bottom), and the second column shows the original memories after a shared backbone and encoder, referred to as $\boldsymbol{M}_t$ and $\boldsymbol{M}_{t+i}$. Obviously, it is hard to distinguish the object from the background on these feature maps. The third column shows the temporal  memory $\boldsymbol{T}_t$ and the spatial  memories $\boldsymbol{S}_{t+i}$. Compared with the original memory $\boldsymbol{M}_t$, it is clear that the $\boldsymbol{T}_t$ has much more attention on the target objects, which indicates that the temporal information does contribute to the distinguishing between foreground and background. The last column shows the temporal-spatial memory $\boldsymbol{E}_t$, $\boldsymbol{R}_t$ after the Residual Gated Correlation and the detection results from top to bottom. Notice the color of the feature map indicates the value. Observing the the original memory  $\boldsymbol{M}_t$ the temporal memory $\boldsymbol{T}_t$ and the final enhanced memory $\boldsymbol{R}_t$, it is easy to find a trend that the values of foreground and background become more easy to separate. The temporal information contributes to recognizing a object by introducing different poses of it. Furthermore, the spatial information helps our \model{} to locate objects with higher confidence score by using spatial transition information. We declare that the reason of such excellent results is the contribution of temporal and spatial information from our TFAM and STAM.

\myparagraph{Results Visualization} We present the results of both the single frame baseline method and our \model{} in Figure \ref{fig:samples}. In particular, the detection results are exhibited in the time order. Compared with the single frame baseline method DETR, Our method shows the priority towards the image deterioration problems. By exploiting the temporal and spatial information, we get a higher confidence score in normal situations and behave much better dealing with occlusion and posture deformation. For example, when the face of a hamster gets occluded by the background, the baseline single frame detector is confused about the category, and easily fooled to predict it as a domestic cat. However, its appearances in context frames are clear and easy to recognize, so our method succeeds in predicting the right category by introducing temporal information. In the second video, the detector is expected to detect several cars and a bus. Interfered by the background and occluded by a car, the baseline method fails at detection in some frames. In contrast, with the help of spatial information, our method can sense the motion of the bus and cars and produce the correct results. In the third video, when two whales get too close, it is hard for the baseline detector to recognize both, causing false detection. In this situation, our \model{} behaves much better according to the temporal-spatial enhancement. It is necessary to introduce temporal-spatial information in this situation to better distinguish one object from another. Consequently, our \model{} achieves much better performance than the single frame baseline method thanks to the temporal-spatial information.

\section{Conclusion}
In this work, we propose a progressive temporal-spatial enhanced transformer towards video object detection. Based on a one-stage object detector DETR, we boost the performance with proper design of introducing progressive feature aggregation. Temporal information and spatial information are proved useful to improve the robustness of detector against image deterioration. We also conduct extensive experiments on the public dataset ImageNet VID to verify the effectiveness of our method. We hope our work can shed light on the research on VOD applying anchor-free approaches.

\noindent\textbf{Acknowledgements.} This work was partly supported by MoE-China Mobile Research Fund Project (MCM20180702), the 111 Project (B07022 and Sheitc No. 150633) and the Shanghai Key Laboratory of Digital Media Processing and Transmissions. And part of this work was done while Han Wang performed as an intern at HIKVISION.

\clearpage
%
%
\bibliographystyle{splncs04}
\bibliography{main}
\end{document}